\documentclass{article}

\usepackage{natbib}
\setcitestyle{square}
\setcitestyle{numbers}
\setcitestyle{comma}

\usepackage{arxiv}
\usepackage{graphicx}

\usepackage[utf8]{inputenc} 
\usepackage[T1]{fontenc}    
\usepackage{hyperref}       
\usepackage{url}            
\usepackage{booktabs}       
\usepackage{amsfonts}       
\usepackage{nicefrac}       
\usepackage{microtype}      
\usepackage{lipsum}
\usepackage{comment}

\title{Hybrid active inference}

\author{
  Andr{\'e} Ofner\\
  Research Focus Cognitive Sciences\\
  University of Potsdam\\
  Potsdam, Germany\\
  \texttt{ofner@uni-potsdam.de} \\
   \And
  Sebastian Stober \\
  Artificial Intelligence Lab\\
  Otto von Guericke University\\
  Magdeburg, Germany\\
  \texttt{stober@ovgu.de} \\
}

\begin{document}
\maketitle

\begin{abstract}
We describe a framework of hybrid cognition by formulating a hybrid cognitive agent that performs hierarchical active inference across a human and a machine part. We suggest that, in addition to enhancing human cognitive functions with an intelligent and adaptive interface, integrated cognitive processing could accelerate emergent properties within artificial intelligence. To establish this, a machine learning part learns to integrate into human cognition by explaining away multi-modal sensory measurements from the environment and physiology simultaneously with the brain signal. With ongoing training, the amount of predictable brain signal increases. This lends the agent the ability to self-supervise on increasingly high levels of cognitive processing in order to further minimize surprise in predicting the brain signal. Furthermore, with increasing level of integration, the access to sensory information about environment and physiology is substituted with access to their representation in the brain. While integrating into a joint embodiment of human and machine, human action and perception are treated as the machine's own. The framework can be implemented with invasive as well as non-invasive sensors for environment, body and brain interfacing. Online and offline training with different machine learning approaches are thinkable. Building on previous research on shared representation learning, we suggest a first implementation leading towards hybrid active inference with non-invasive brain interfacing and state of the art probabilistic deep learning methods. We further discuss how implementation might have effect on the meta-cognitive abilities of the described agent and suggest that with adequate implementation the machine part can continue to executed and build upon the learned cognitive processes autonomously.
\end{abstract}

\keywords{Active Inference \and Hybrid cognition \and Artificial Intelligence}

\section{Introduction}

The last decades have shown increased transfer of insights from the fields of neuroscience and psychology into machine learning systems and vice versa. State of the art machine learning systems are often based on artificial neural networks and make use of mechanisms inspired by human cognitive abilities on various levels, such as hierarchical processing, attention or memory mechanisms \cite{rybkin2018unsupervised, wayne2018unsupervised}. Despite the existence of a large variety of complex approaches to network design and training, such as training agents with reinforcement or meta-learning strategies, there is still a lack in artificially intelligent systems that are able to generalize and show human-like behaviour on higher cognitive levels \cite{blanchard2018neurobiological}. 

Recently, the performance of deep artificial neural network models has been improved by comparing internal activation to the human brain signal for shared tasks. Furthermore, researchers have started to systematically compare machine and human cognition not only on sensory processing or classification tasks, but for more complex behaviours, especially for game-like tasks. However, this is still largely done with a significant information gap between human and machine. 

In turn, within the field of neuroscience, machine learning techniques such as deep neural networks are a driving force for understanding brain signal. Here we argue that establishing a close connection between human and machine within a unified cognitive system can lead to emerging behavioural properties within a machine learning sub-system. We argue that through comparative learning of sensory environment and physiological data of the human 'host' a machine learning agent can learn to adapt the human brain and physiology into its own processing. 

As a theoretical basis, we rely on the active inference account \cite{friston2009free} decision making. This normative account is based on the principle of free energy minimization, that is maximizing model evidence or minimizing surprise. Following the active inference account, a single objective function leads to complex behaviour and emergence of various cognitive processes, such as exploration. The active inference account is widely accepted within the field of neuroscience and accounts for a large variety of phenomena on a broad spectrum of cognitive levels. Furthermore, aspects of this theoretical framework has recently been implemented into a variety of machine learning models, such as deep neural network based models of predictive coding or active inference \cite{zhong2018afa, ueltzhoffer2017deep}. Here, we formulate a hybrid cognitive system that integrates the hierarchical active inference process hypothesized to be executed by humans into a larger active inference process within a hybrid cognitive system. This approach is inspired by the observation that hierarchical processing within humans allow for the integration of complex cognitive processes into even more complex ones, e.g. a variety of subconscious processes that are driven by conscious processing. We suggest that in such a system, self supervision on increasingly high cognitive levels can arise, as increased understanding of human cognitive processing can be turning into self-supervision for more complex cognitive processes. Each of these processes in turn establish the next level within a shared hierarchical processing. We suggest that, using this approach, meaningful representations of the brain signal and the environment can be learned in an unsupervised way. Furthermore, we argue that, starting from a certain cognitive level, the self-supervision process can be augmented by bi-directional human-machine interaction along the levels of the learned hierarchy.

The remaining sections of this article are structured as follows: The first section provides an overview of the research and terminology used in the emerging field of hybrid cognitive systems. A further section briefly reviews the state of the art methods that combine machine learning and neuroimaging to advance research in both fields. The subsequent sections briefly summarize the active inference account and its implementations, which serve as the foundation to hybrid active inference. We then give an example for an implementation using electroencephalography (EEG) and a deep neural network based implementation of active inference based on our previous research on shared representation learning. A final section discusses the presented framework in regard to emergence, artificial general intelligence, agency and meta-cognitive properties.

\section{Hybrid cognitive systems}

Research on cognitive processes that combine living and artificial systems is a nascent field and still lacks precise terminology. Recently, the term "neurobiohybrid" has been coined to summarize the cases where a neuronal processing system, such as the human brain, is coupled with a biomimetic artifical system. Thus, "neurobiohybrid" are a subclass of "biohybrids" that show "close physical interactions at the molecular, cellular, or systems level, eventually leading to information flow and processing in one or both directions" \cite{vassanelli2016trends}. Often times, such hybrid cognitive systems are built with the intention to restore or augment human brain function and thus are focused on understanding the human in the loop \cite{vassanelli2016trends}. Many of these systems show a tight physical and biological coupling through invasive interfacing with the brain. However, the term "neurobiohybrid" accounts for non-invasive interfacing as well, even if they large provide uni-directional feedback between living and artificial subsystem \cite{vassanelli2016trends}. Problematically, as "biohybrid" systems are described as "establish[ing] physical interactions with information exchange" here we opt to place our agent within the looser term 'hybrid cognition' and refer with it to all cognitive processing that involve two functionally seperatable subsystems, no matter if they are physically coupled.  

\section{Previous work combining neuroimaging and machine learning}

\subsection{Using human brain activity to enhance machine learning performance}

Several recent studies have started to exploit the human brain as a means to evaluate, guide and improve state of the art machine learning models, especially within the domain of deep learning. For example, for the case of a deep predictive coding network performing next frame prediction on images from videos, Scheirer et al. have demonstrated "that models with higher human-model similarity are more likely to generalize to cross-domain tasks". For this, they suggest a "human-model similarity metric" that is "calculated by measuring the similarity between human brain fMRI activations and predictive coding network activations over a shared set of stimuli" \cite{blanchard2018neurobiological}. Earlier, Cox et al. have proposed a "neurally-weighted" approach to machine learning. They provided an examplatory case where fMRI measurements of the human brain are used within artificial neural network training. The resulting "neurally-weighted classifiers are able to classify images without requiring any additional neural data" \cite{fong2018using}.

\subsection{Using machine learning models to model and predict the human brain}

A variety of studies exist that employ machine learning to model and predict aspects of human cognitive processing. Once again, many of these studies employ deep artificial neural networks \cite{wen2017neural, wen2018transferring}. Within this research area, two common approaches are used to test the explanatory power of a predictive model: Encoding models that predict patterns of brain activiation expressed, for example as fMRI or EEG measurements. A second class are decoding models, which transfer a stream of brain signal into the domain of stimuli or behavioral patterns which cause the activations \cite{naselaris2011encoding}.

The range of predicted cognitive processes is broad. However, a substantial amount of these studies focuses on the visual processing by comparing the response of deep convolutional neural networks (CNN) to visual stimuli to the human brain recorded with functional magnetic resonance imaging (fMRI) \cite{wen2017neural, wen2018transferring, qiao2018accurate}. For example, Wen et al. showed the possibilty to use CNNs to both encode and decode fMRI measurements of humans watching natural movies. In their study, a CNN encoding fMRI data, i.e. synthesis of fMRI signal based on the sensory input with the model, could produce meaningful patterns of cortical activation regarding category representation, contrast, and selectivity of the presented stimuli. A separately trained CNN decoding model enabled visual and semantic reconstruction of the stimuli from fMRI data \cite{wen2017neural}.
Other studies demonstrate that functional and representational similarities between deep learning models and hierarchical visual processing in the brain can be used to deepen our understanding of how the brain works \cite{greene2018shared, cichy2017dynamics}. 

But encoding and decoding is not limited to visual processing. Multiple studies model and predict and classify neural correlates, from the behavior of single neurons to predicting and classifying psychiatric and neurological disorders \cite{vieira2017using, dvornek2018combining, heinzle2018dynamic}. A substantial amount of these studies are methodically based on dynamic causal modeling of neuronal activation and functional connectivity in the brain \cite{marreiros2010dynamic}. These models account for information expressed in a single modality, such as fMRI, MEG or EEG, or across multiple modalities \cite{bonstrup2016dynamic, kiebel2008dynamic, grana2017dynamic, david2006dynamic}. Recently, there has been a shift towards employing deep learning for approaches using dynamic causal modeling \cite{wang2018generalized}. 

Even though, strictly speaking, they were not designed from a machine-learning perspective, models based on the active inference account have been used to model and predict a vast range of human behaviour and brain activity. The active inference account will be discussed in depth later in this article. For example, these hierarchical generative models under Markov decision processes were used to model decision making under uncertainty, saccadic eye movements, reading behavior as well as to model and predict electrophysiological brain responses, such as P300 activations or the phenomenon of mismatch negativity \cite{friston2018deep, mirza2016scene}. Recently, an implementation of an active inference based reinforcement learning agent within the OpenAI Gym environment was suggested as a "principled algorithmic and neurobiological framework for testing hypotheses in psychiatric illness" \cite{cullen2018active}.

\subsection{Representations shared in the human brain and machine learning models}

Although implicitly done in some of the previously discussed studies, a range of research focuses explicitly on learning representations that are shared in the human brain and machine learning models \cite{du2017sharing, shen2017deep, ofnershared, richard2018optimizing}. These studies typically make use of multi-modal deep neural network models that learn to map features found in the stimuli as well as the neuroimaging modality into a shared latent space. The latent space is then used to reconstruct brain signal and stimuli simultaneously. This type of explicit shared representation learning has been successfully done for various neuroimaging modalities, such as fMRI and M/EEG \cite{shen2017deep, ofnershared}.

\section{Active inference and predictive coding}

The active inference account was coined by Karl Friston and refers to a view on cognition that explains all behavior as the result of the minimization of a single function, namely that of "free energy". Generally speaking, in its role as a predictive processing theory it explains not only action, perception and inference in an organism, but also accounts for the emergence of intentions and beliefs. Predictive processing has even been mentioned as a general principle for all psychological phenomena \cite{kirchhoff2017predictive}. 

\section{Learning generative models with free energy minimization}
Based on the Information Free-Energy (IFE) principle, the active inference account suggests that every agent has to maintain a stable internal structure despite having to deal with a complex and dynamic environment. Generally, the agent is embodied and embedded into a complex environment.
In order to survive, every agent must minimize the entropy of its internal states which are influenced by 'surprise' stemming from sensory data. This is one of the core aspects of active inference: As agents do not have access to the underlying causes (or states) of their environment, they must optimize their sensory processing to minimize surprise. Surprise here does not refer to surprise in its psychological sense, but rather to a quantity describing the difference between predicted and actual sensory data. 
Sensory data here stands not only for the perceptions from the agent's environment, but is used in a generalized fashion to refer to any sensory activation external or internal to the agent.
An important aspect of the active inference account is that behavior arises from the constant process of the agent actively minimizing the discrepancy between its predictions and the sensations. This means, that through internal generative models of its sensors and actuators and its environment, the agent acts upon the world in order to minimize surprise and essentially make the predictions true. This means that driven only by the process of entropy minimization, active exploration of the environment and with that, the bounds of the agent's embodiment and cognitive capacity arises. 
Here, the ambiguous relation between intention and prediction surfaces: The agent always has a variety of options how to deal with surprise, e.g. continuing with an intended action or using the resulting sensory data to update its predictions about the future. This means that with continuation of the active inference process, increasingly complex intentions and beliefs emerge as increasingly complex and accurate generative models are learned. 
Friston suggests that during this process an agent creates a generative model of its environment and uses this general principle to account for a large variety of phenomena. Ongoing research continuous to explain a variety of psychological phenomena with active inference, for example by describing emotion as a process of active inference on interoceptive sensations \cite{seth2013interoceptive}. 

\section{Action and perception in active inference}
It is important to note that the active inference account casts action and perception as mutually dependent processes that arise implicitly from free energy minimization and are executed in a closed loop. Under this view, action serves to minimize surprise by changing the input to an organism's sensors. The surprise stemming from this input, in turn, is minimized by adapting predictions by perception \cite{friston2009free}. When designing an agent that performs active inference, it is thus crucial to model action and perception as very closely related processes, working jointly within each sensory modality. These insights lead to the relation between active inference and the predictive coding theory in neuroscience, which has been at the core of many recent studies in the field. The predictive coding theory suggests that the sensory system is not a one-directional hierarchy of feature extractors working on input data. Instead, the the theory suggests that the brain processes inputs rather in terms of a discrepancy between predictions and input data. Here, the resulting prediction error is processed and predicted across a hierarchy of sensory regions that are mutually informative \cite{adams2015active, kok2015predictive}.
This means, that multiple regions process the expectancy of what other linked regions are expecting, leading to a deep hierarchical architecture. Friston argues that hierarchical predictive coding, which is line with many observations about neural processing in the brain, can be used to implement active inference \cite{friston2008hierarchical}.  
A core idea in the resulting hierarchical predictive coding model of active inference is that action (i.e. motor signals) are processed in analogy to perceptual signals and therefore both types of signal underlie a joint type of processing. It should be noted, that from this view, predictive coding is resulting from the active inference process and its existence is thus explained by the account \cite{friston2008hierarchical}. From a neural perspective, it has been argued that this principle is implemented in the brain by neuron populations that encode prediction and prediction errors. In this process, involved neurons also account for the precision the prediction, which is used to explain the emergence of attentive and memory processes \cite{friston2009free}.

\section{Active inference and the Bayesian brain}
From a mathematical perspective, active inference is commonly based on the probabilistic process of Bayesian inference. The difference of free energy and surprise, i.e. the divergence between the representations of sensory data in the agent and its true causes, is expressed as the divergence between two probabilistic distributions \cite{friston2009free}. The probabilistic representation \cite{friston2009free}. Speaking more intuitively, these probabilistic representations can be viewed as internalized generative models of the agents body and its environment. Depending on the complexity of the organism, it has been argued that these generative models are used in various levels of cognition. For example, a human could use such a generative model to predict (or extrapolate) specific aspects about future states of his or her environment, for example when imagining how a car might look from a side that is currently hidden to him or her. Here, based on previous experience, the resulting generative model of a car lends the agent to efficiently predict its environment and deepen its understanding through the resulting possibility to verify the models correctness. It should be noted that these generative models account for both the external and internal to an agent, i.e. extero-, proprio-, and interoception. 

It has been argued that the complexity of the hierarchical interconnections is largely dependent on the embodiment and ecological niche, i.e. surrounding, of an agent \cite{linson2018active, hasson2008hierarchy}. For example, the depth and complexity of generative models in bacteria is simpler than in humans, where a hierarchy of nested regions allows spatiotemporal integration along multiple time-scales. Creating generative models that work on larger time scales while being embedded into a hierarchy is suggested to allow humans, for example, to integrate "syllables into words, words into sentences, and sentences into a narrative" \cite{linson2018active}. In summary, the active inference account can be described as a variational principle of least free energy, namely Bayesian inference in the brain, which causes the emergence of a hierarchy of increasingly complex generative models and cognitive processes.

\subsection{Machine learning implementations of active inference and predictive coding}

A large variety of implementations inspired by the active inference account as well as the predictive coding theory has been designed and evaluated in recent years \cite{ueltzhoffer2017deep, mcgregor2015minimal, zhong2018afa, han2018deep, wen2018deep, dora2018deep}. The original implementations by Friston and colleagues were note designed from a machine learning perspective, but rather to model and validate implications from the active inference account. These implementations are based on a hierarchical architecture of nested processing units acting as Markov decision processes \cite{friston2018deep}. However, the core ideas of the active inference account have recently been transferred into the realm of state of the art machine learning, most notably within the domain of probabilistic deep learning \cite{ueltzhoffer2017deep, mcgregor2015minimal, dauce2017toward}. These models exploit the similarity of the guiding principle of active inference, namely the minimization of variational free energy, to the variational free energy that serves as the objective
function used in the variational auto-encoder (VAE) \cite{kingma2013auto}. VAEs are an established type of deep generative modeling within the deep learning community. 
Several slightly different implementations of deep learning based predictive coding networks exist, where each focuses on a selection of aspects of the theory while ignoring other. For example, Cox et al. suggested the "PredNet", an approach that employs prediction error propagation in deep convolutional neural networks \cite{lotter2016deep} and is applied to video frame prediction. It has recently been augmented to account for action modulation of the perceptive processing \cite{zhong2018afa}. Other versions focus on employing similar networks for object recognition \cite{wen2018deep} or local recurrent processing to enhance model performance \cite{han2018deep}. It should be noted, that these models are generally deterministic in nature, but probabilistic versions have been suggested \cite{lotter2016deep}. 

\section{Hybrid active inference}
Inspired by the increasingly detailed descriptions active inference as a mechanism underlying human cognition and its successful application to state-of-the-art machine learning, we suggest that the time is ripe to directly fuse both domains and leverage hybrid cognitive systems. In this section we formulate our view of a hybrid active inference agent and give an overview of possible implications of such a system on human and artificial cognition. We then continue to describe how such a system might be implemented with state of the art machine learning and available sensors for brain interfacing.

Based on the existing terminology for neurobiohybrids, we define a hybrid active inference agent to be a cognitive agent, that:\newline

a) consists of at least one natural (human) and one artificial part \newline
b) is internally coupled through a shared embodiment and embedding into the same environment \newline
c) implements active inference as the main mechanism to process information in both sub-systems \newline
d) allows both active inference processes to inform each other and share cognitive resources through bi-directional feedback \newline

The resulting cognitive agent shows a hierarchical structure that integrates processes from artificial and human cognition between various levels. Figure 1. shows a schematic overview of the multitudes of connections between natural and artificial processes that are included in the hybrid cognitive agent. The amount and type of connections between artificial and natural processes is highly dependent on the cognitive level of the artificial part and the available interface. 

\begin{figure}
\includegraphics[width=\columnwidth]{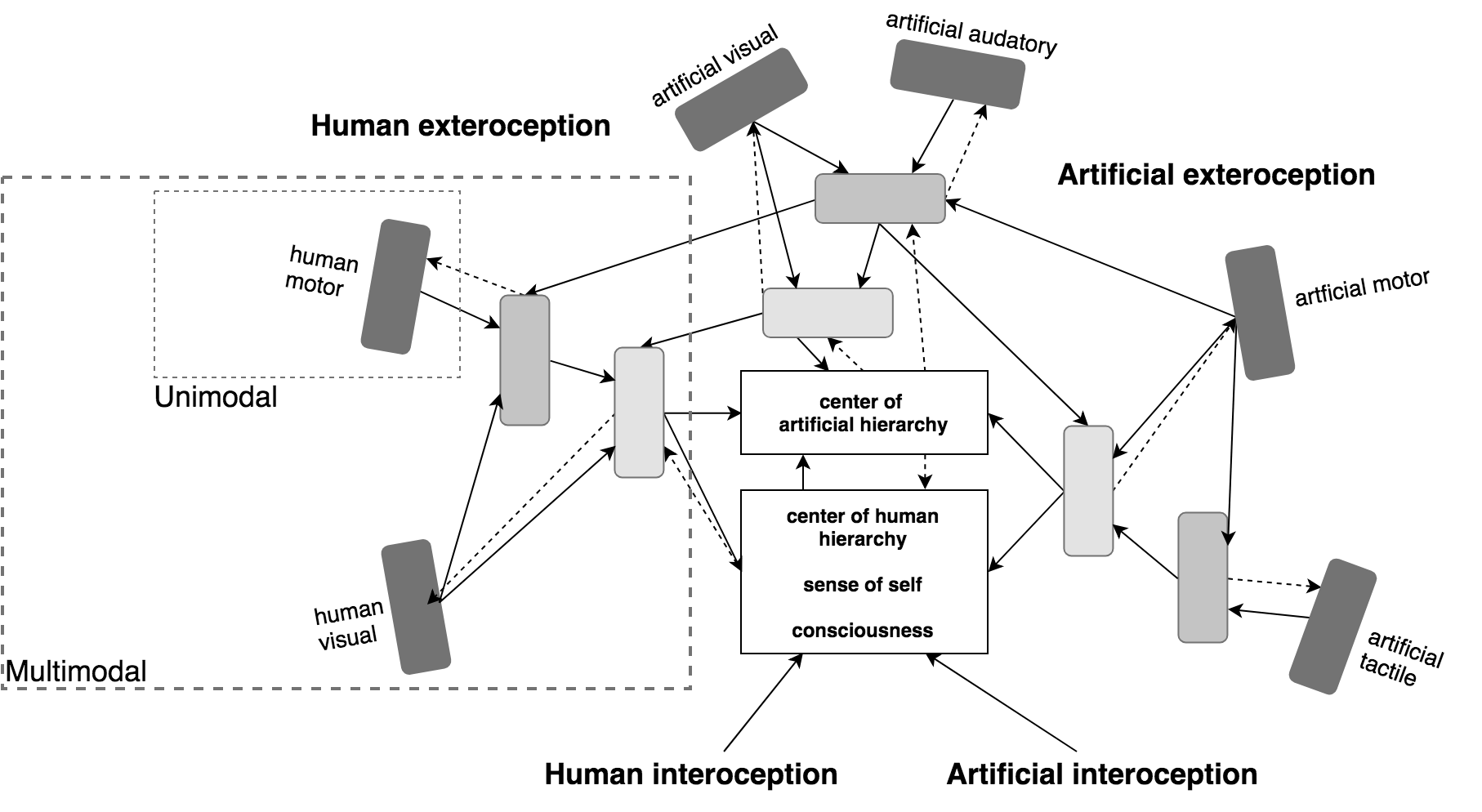}
\caption{Schematic illustration of the hierarchical structures within a hybrid active inference agent. Continuos lines indicate examples for forward connections, while dashed lines indicate backward or lateral connections. The example shows an agent that has already established feedback connections between areas of high cognitive level, such as access of human conscious representations in the artificial active inference hierarchy. In earlier stages of integration, the agents internal structure is dominated more by unimodal connections.}
\label{fig:architecture}
\end{figure}

\subsection{Surprise minimization as a means to canonical processing}
The basic idea underlying the hybrid active inference agent is that free energy (or surprise) minimization acts as a driving force that forces the (probabilistic) representations within the artificial part to efficiently explain as much as possible of its sensory inputs. So far, this process is identical to the workings of an artificial active inference agent. However, through defining the neuronal activations to be an additional input, we define a system that tries not only to model the generative causes of its own environment, but additionally those that cause the brain signal. As active inference is thought the be the basic principle for the neural activations in the human, we hereby effectively model the active inference hierarchy of the artificial part to 'sit on top' of the hierarchical processing in the human.
The resulting unified system now features two hierarchies of predictive processing with access to the same sensory environment. 
As the machine part constantly tries to minimize surprise on its sensors, it must learn meaningful representations about the incoming data within lower levels of the inference hierarchy. In order to explain the multi-modal inputs, the emergence of more complex cross-modal generative models in deeper layers of the hierarchy helps to improve the overall precision. Within the deeper layers, the learned representations must explain the brain and the environment simultaneously. Given that, following this principle, the agent has learned meaningful representations for several modalities, it now has (still governed only by the free energy minimization) the possibility to transfer the outcomes of predictions between modalities. With that, it will use representations derived from the environment to help explain the brain response and vice versa. It should be noted, that we required the natural and artificial sub-system to have a shared embodiment. This means that both entities are embedded into the same environment and stimuli that are processed by the artificial part are also processed by the brain.

\subsection{Unsupervised active inference}
It has been noted that one of the most fundamental problems in designing a machine learning version of active inference is the lack of description of what the artificial agent "should expect to experience" \cite{mcgregor2015minimal}. This refers to the issue that the type of these expectances themselves are not specified by the active inference account, the theory 'only' gives the principles under which they emerge. By casting artificial active inference within a hybrid cognition setting, we hope to at least reduce if not eradicate this sort of problem. State of the art machine learning and robotics often times resorts to supervised or reinforcement learning in order to create intelligent and emergent processing. Within our formulation of a hybrid cognitive agent however, there is arguably no need to specifically implement expectancy's, evaluating the existence of those or resorting to supervised or reinforcement learning. Instead, we suggest that the purely unsupervised task of minimization of surprise on the brain signal is enough to start an active inference process leading to emergence of intelligent behaviour.
At the core of this suggestion is, that the artificial part in the hybrid system is fused with the expectances (and with that, beliefs, intentions, actions, memory and so on) of the natural part. We argue that, while this removes the need to supervise them explicitly, the artificial agent still has the possibility to shape the entirety of its own inference process, which has to work as efficiently as possible within the overall system. 

\subsection{Bi-directional feedback between human and machine}
However, this does not mean that the hybrid cognitive agent cannot use supervised or reinforcement learning efficiently integrate the artificial part into the overall system. On the contrary - once the machine part has reached certain abilities, such as generating meaningful feedback from what it understands of the brain signal, this might turn into a driving force for the active inference process. If external feedback between the involved human and machine is enabled through supervised or (interactive) reinforcement learning, it here thus serves as an efficient means of information exchange between both active inference processes. This assumption is based on the aspect that both the causes as well as the consequences of the supervision remain processed within the bounds of the unified cognitive system. For many implementations of the hybrid agent, this sort of feedback might be preferred, as it does not require invasive sensors and is compliant with state of the art training methods for existing machine learning models.

\subsection{Quantity and quality of feedback as an emerging property}
At early stages of model training, there will be little meaningful feedback possible between the two involved sub-parts. One could argue that this resembles a large amount of current brain-computer interfaces, which provide only unidirectional or very little bi-directional feedback. However, for the described hybrid agent, we require the possibility of access towards feedback between the two active inference processes rather than one that is constantly and used by the system. This is based firstly on the assumption that the type of feedback will be shaped dramatically by the agents embodiment, for example by the type and amount of sensors (such as non-invasive versus invasive brain interfacing). Secondly, it is based on the assumption that feedback is an emergent property within the hybrid agent and thus is not required to be executed at all time.
There are many types of emergent feedback, that might arise at different levels of integration. For example, the artificial part might eventually integrate into language processing and enable structured access to thought content, which can be used for verbal communication within the system. Furthermore, it might be possible to trace the content of conscious processes, especially with regard to the content of working memory. This could be imagined both as a possibility for feedback as well as meta-cognitive processing on top of conscious representations. It should be noted, that consciousness here largely refers to its interpretation as globally integrated awareness.

\subsection{Variable amounts of sensors and actuators}
Closely related to the issue of potential for feedback is the quantity and quality of the sensors that are providing input to the active inference process in the machine part. The presented formulation of hybrid cognition strictly requires access to the human brain, or rather the human active inference process. The amount of additional sensors connected to the human body and the environment however, is arbitrary. This is based on the hypothesis, that a well integrated machine part can use representations of these domains in the brain rather than inferring them from ground up through its own inference process. In practical systems however, the quality of access to brain data might fluctuate heavily, be of global nature rather than local and be very noisy. This is especially the case for low-cost and non-invasive brain-interfaces, such as mobile EEG systems. In detail, while the machine part might have bi-directional access to local neuron populations with high temporal and spatial resolution in invasive interfaces, in many other cases this might not be possible. In such cases, the artificial part can 'offload' less onto fine-grained access to representations in the brain and needs to work more directly on sensory inputs. In summary, this means that it might be possible to compensate low-quality access to neurophysiological signal with large access to multi-modal sensory inputs.

Following the idea of grounded cognition, we suggest, that the amount of sensors required for active inference decreases with increasing understanding of the brain signal. This thought is based on the idea that once the machine part has access to the hierarchy of representations inside the brain, they can be used directly, as they have been grounded within the human part before. We suggest that the process of (partially) relying on grounded representations in the brain rather than learning all through the machine's own inference process might significantly accelerate the learning process.

\subsection{Autonomy}
Another key point of our formulation of a hybrid cognitive agent refers to the degree of autonomy of the machine part. To avoid confusion - here we refer to autonomy as the machine's ability to infer on the artificial sensors rather on the human brain. As we speak about shared cognitive processing and the machine part has (theoretically) access to all processing in the brain, this approach hybrid cognition does not automatically imply that the machine agent learns to completely mirror all brain function. Instead, for any sort of cognitive function, we expect it to mirror the brain only to the point where it can rely on its understanding of the brain to achieve better predictions. This implies, that the sort of cognitive functions that are mirrored versus those that are 'borrowed' from the brain is highly dependent from the structural design of the artificial part. For example, an agent with access only to information about bodily movements of a human (or access to its own actuators), might learn to mirror and integrate aspects of motor control before anything else. Another implementation of our described hybrid cognitive agent might first rely more on mirroring aspects of visual predictive processing. 

\subsection{Exteroception and interoception}
It has been argued that, next to processing the environment as well as learning to understand the physical bounds of one's body, access to the body's internal states plays a crucial role for complex behavior to arise in humans. Furthermore, processing of internal signals has been demonstrated to suppress predictive processing within other domains \cite{seth2013interoceptive, seth2016active}. For this reason, 
we suggest to give the artificial part access not only to sensory experience from the environment, but also from the human's body and movement as well as physiological state. However, in congruence with the section on sensors and actuators, information from these sensors might also be extracted implicitly from the brain rather than from the measurements, as they are grounded within the hybrid system though the natural part. As we are basing the hybrid agent entirely on the active inference principle, action and perception are treated within the artificial part identically as in the human part, namely to be closely related to each other and working in a loop to create predictions with high precision. It should be noted that the existence of efference copies and corollary discharge of motor command has been demonstrated in the human brain. These copies of action commands sent to the muscles are distributed in the brain and alter other processes \cite{feinberg1978efference}. Provided sufficient brain interfacing sensor quality, the artifical part could use theses activations rather than sensory information.

\subsection{(Virtual) hybrid embodiment}
We explicitly require the hybrid agent to be jointly embodied with the human part. Embodiment here should be viewed as the sum of all information that is deduced from physical information accessible to active inference within human and machine. This could either happen directly or by using active inference on the respective other part. This means that, at least theoretically, both agents must have access to knowledge about the entirety of physical bounds of the overall cognitive system. In line with research on embodied intelligence, we suggest that such stressing of a continuously definable embodiment is a crucial aspect for meaningful behavior to arise within the active inference process of the machine. 
However, there are many types of implementation conceivable for how such a unified embodiment is realized and research suggests that embodiment is a highly complex issue. For example, one might design a system with a virtual reality component. While this might sound like adding complexity at first, we suggest that this might actually accelerate the hybrid agent to successfully predict based on its shared embodiment. This is based on the idea that virtual environments provide simpler sensory inputs and thus provide less complexity on defining what is part of the body. While one might argue that this might not resemble 'real' embodiment enough. However, we note that in such situation the human is still embedded into the real world. Such an embodiment therefore always spans both domains, while possibly enabling more efficient inference. Such a setting might be especially useful for early stages of training.

\section {Hybrid active inference with non-invasive brain interfacing and deep generative models}

\subsection{Previous work on deep generative models in non-invasive brain interfacing}
Using generative machine learning models, especially hierarchical variational (bayesian) models, in combination with deep neural networks is already an established method within the fields of EEG source localization, classification and stimulus reconstruction \cite{lucka2012hierarchical, wu2011hierarchical, wipf2009unified, ofnershared, kavasidis2017brain2image}. Many of these approaches have been implemented in the context of non-invasive brain-computer interfaces. Often times, these enable the execution of mental commands or visualize learned representations \cite{kavasidis2017brain2image, ofnershared}. In recent years, these interfaces have been made more efficient by integrating multiple neuro-physiological measurements such as combinations of EEG and MEG sensors with fNIR and fMRI \cite{hong2018feature, corsi2018integrating, zotev2018real} or implementing context-aware systems \cite{ozdenizci2018hierarchical}. Often times, feedback from or to the machine learning system is a core part of the interface and a large diversity of non-invasive feedback methods has been developed and tested \cite{zotev2018real}. These range from 'neurofeedback', i.e. feedback about (conscious) alteration of specific brain activation, over motoric control of artificial actuators or robots to tactile, auditory or visual feedback \cite{zotev2018real, angelakis2007eeg, lukoyanov2018efficiency, millan2004noninvasive}.
Generally, these methods follow a 'traditional' view on machine learning and strictly distinguish between the human as a source of data and a machine learning model, which denoises the signal and learns useful representations from short snippets of the EEG data stream or across longer temporal sequences \cite{lucka2012hierarchical, wu2011hierarchical}. However, recently suggestions to implement more human-centered interfaces were made \cite{lotte2018bci}.

\subsection{From multimodal non-invasive interfacing to hybrid active inference}

Transferring the principles of predictive coding and active inference into artificial systems is not trivial. Even for the domain of neural processing, the exact mechanisms are active debate. However, all of the currently existing implementations inspired by these theories have provided promising behavior and there are already a wide range of studies that employ the discussed processes to advance problem solving with state of the art deep learning and autonomous agents \cite{cullen2018active, wen2018deep, wayne2018unsupervised}. Here, we briefly summarize our previous work on non-invasive multi-modal brain interfaces and give suggestions how to turn them into a basis for active inference.

In our previous work designed a framework for shared representation learning from multiple inputs, projecting brain activation (expressed in EEG signal) environmental stimuli into a joint latent representational space. For this, we used the possibility of unsupervised learning of auditory stimuli and their associated brain states with multi-modal auto-encoding VAEs \cite{ofnershared}. Here, we employed deep convolutional neural networks to process raw and pre-processed EEG signal as well as audio signal before projecting them into the VAEs shared latent space. The model is trained in a straightforward way by computing the mean squared error between reconstructions and targets for each modality. We further have demonstrated the ability to use the model to interpolate and thus introspect the shared latent space. This framework (based on deep variational canonical correlation analysis or VCCA) was designed to process inputs from arbitrary amounts of sources. While we highly welcome approaches based on a different approach, we suggest to enhance this idea of multi-modal data fusion with a variety of augmentations in order to create a basis for the previously described hybrid active inference: \newline

a) Drawing from the recent advances in designing deep learning based versions of the predictive coding and active inference process, we suggest to exchange the feed-forward encoders in the model with those that implement predictive coding or active inference. Here, it might be a sensible approach to start with predictive coding based encoders and iteratively increase model complexity until the unified active inference framework is reached. This way, representations learned purely from data compression, multi-modal data fusion and predictive coding can be analyzed and compared with representations in more comprehensive models. A first useful step towards hierarchically organized inference might be replacing each of the feed-forward encoders with a deep predictive coding model, such as implemented by Lotter et al. \cite{lotter2016deep}. This gives the model the ability to perform modality specific predictive processing, while the overall outcome of comparing predicted sensory data with the actual inputs is propagated into the latent space which integrates all modalities. \newline

b) Many of the existing network designs implement only certain aspects in depth, while ignoring other aspects covered by the predictive processing theory. For example, PredNet accounts for a computation of sensory inputs inspired by the predictive coding theory, but does not (yet) feature any probabilistic parts. We suggest that, once again, it might be a sensible approach to evaluate the performance of simple models and then adding more complexity.

c) In order to account for sensory signals subject to human extero-, proprio- and interoception, we suggest to implement encoders in the model which are connected to the human's environment and accessible physiological signals in addition to the brain signal encoder. \newline 

d) We ultimately expect the agent to rely on and integrate the corollary discharge of motor commands as its own motor behavior. However, we assume this to be a very complex task which might be reached only with very complex models and maybe even only with invasive brain interfacing. To deal with this problem, we suggest to explicitly allow for action modeling in early stages of development. For example, one could use one or several modalities that monitor the movement of environment or human body which is used for modulating the predictions made in perceptual modalities. Such a network, based on action modulation of a deep predictive coding network, has been recently been implemented by Zhong et al. \cite{zhong2018afa}.

e) Active inference is described as a process that is hierarchical in nature, with the regions along the hierarchy being in constant exchange. A first step might this be to create hierarchically organized versions of the active inference (or predictive coding) based encoders. Exchanging information between the sub-systems might be achieved with vertical and horizontal cross-connections, possible after re-iterated compression into lower-dimensional latent representations. \newline

f) Recently, Yoshua Bengio suggested to use low-dimensional representations summarizing more complex underlying processes in order to create complex behavior and possibly even consciousness in machine learning systems \cite{bengio2017consciousness}. Here, we suggest to adapt this strategy to hybrid representations, i.e. create a hierarchy of increasingly low-dimensional latent representations embedding within the model. 

g) There are many ways to allow the described model to process temporal aspects of the input, for example by using local or global recurrence or by allowing the model access to episodic memory. Furthermore, using recurrent and memory-based approaches might further increase the representational capacity of the discussed model. However, recurrence and memory is generally accompanied by significantly increased processing time \cite{han2018deep}. It should be noted, however, that within the hybrid cognition framing, the machine learning agent already has access to memory-based human cognitive processes. Thus it might be useful to specifically design memory resources either for sensory input or the modality capturing the brain signal.

\subsection{Training the hybrid active inference system}

\subsection{Pre-training}
We suggest that there are several different approaches to training the described non-invasive deep hybrid active inference agent, each of which has its own advantages and disadvantages. Generally speaking, it might be useful to distinguish between a pre-training and training stage. Within the pre-training stage, low-level representations, for example for visual processing, could be learned by using large amounts of brain signal recorded simultaneously with data from physiology and the environment. Using this approach, the training stage could work on already established representations within lower layers and learn to re-use them for more complex ones. We suggest that this approach is useful especially in combination with the very noise nature of EEG signal. Furthermore, this allows for training with large amounts of data before attempting any real-time execution. Access to large quantities of data is commonly regarded as a crucial factor for the efficiency of deep learning systems. Additionally, temporally separate pre-training could allow to include sensor types that are not available during real-time execution, such as invasive sensors or expensive machinery, such as used to record fMRI data. Using the ability of the model to integrate multiple modalities into cross-modal representations provides a useful way to integrate high quality measurements into the training process. We suggest that this might go in line with the theory of grounded cognition, where high level processes do not need access to fine-grained sensory data every time a derived concept is re-used.

\subsection{Reinforcement learning and feedback}
Recently, much progress has been made in improving reinforcement learning strategies. We suggest that, especially for real-time execution, a reinforcement strategy might be useful. Recently, interactive reinforcement learning has been proposed and evaluated in various research projects across machine learning and robotics. The combination of feedback mechanisms with reinforcement learning, i.e. supervision by the human part within the hybrid system, can serve as a means to establish and enhance associate learning between human and machine part. Crucially, within the proposed hybrid agent, the brain signal can viewed as both the training data as well as the reward signal. Especially for more advanced models and to provide the means for experimental tests on the described system, it might be useful to visualize the internal generative model(s). We suggest that, next to introspecting the learned generative world models, one could employ these modules as a means of human-machine feedback. We think that actively using these visualizations might be useful, for example, for interactive reinforcement learning \cite{thomaz2005real}. By constantly monitoring the output of a generative model, it might be possible to automatically supervise for emergent properties. For example, while jointly reading a sentence about a car, one of the learned generative models might visualize a picture of a car, inferred either from the brain or from the machine's own sensory input. In this example, we suggest that the meaningfulness and quality of visualization will be noticed by the human and the resulting brain response can in turn be processed by the machine part. 
We do not require the implementation of externalized feedback mechanisms to fulfil the requirements of hybrid active inference. However, a large variety of possible feedback mechanisms for shared cognitive acts are conceivable, such as letting the machine part control a physical device. We think that such an approach stresses the ability of the system to enhance human capacity, leading increased associative learning. Furthermore, it might be sensible to aid the learning process by monitoring the machine's ability to process physiological feedback signals, such as emotion-based facial responses or specific activation's in the brain related to valence. We suggest that such approaches could improve the capacity of state of the art reinforcement learning agents, even if they are trained with a focus on autonomous execution.

\subsubsection{Towards artificial general intelligence}
While the cognitive capacity of implemented hybrid active inference agents is hard to predict and heavily influenced by current and future developments, especially within machine learning and neuroengineering, there are several aspects that hint at their potential: Firstly, by having access to human thought on all hierarchical levels, the artificial part is equipped with a large potential for meta-cognitive acts across many levels of complexity. Arguably, in the described hybrid system, the content of conscious thought is part of what is processed by the artificial part. Intuitively, having access to intelligent, adaptive interfaces that can learn to predict content of conscious thought can have a big impact on human cognition.
Secondly, our formulation of hybrid cognition allows (or even forces) the agent to 'remain' largely autonomous, as the process of inferring on sensory stimuli is happening simultaneously with inference on the brain, but is not designed to only explain the brain. This means that, if desired, one could separate the two sub-systems at any time and exploit what has been learned in the artificial part for autonomous behaviour. Within the context of biohybrid systems, it has been suggested that the process of "seeding" an artificial agent with data from humans might provide a basis for a high level of cognitive capacity. This is in line with the previously results from studies that have used projections of stimuli into associated brain states as a means to enhance the capacity of deep learning models. While this happened in a uni-directional process, we suggest that in a closed loop this effect might increase.

\section{Current stage of implementation and perspective}

We provided an overview on the growing connections between machine learning and neuroscience. We suggest a integrative approach to enhancing human cognitive functions while advancing towards general artificial intelligence by formulating a hybrid cognitive agent that is based on the basic principle of surprise minimisation. We are currently running tests with different versions of the described framework, featuring deep predictive coding encoders in combination with variational inference. We test hierarchically organised architectures with decreasing dimensionality of latent spaces and increasingly large time-scale towards deeper layers. In order to provide a basis for other researchers to design more advanced models and enable meaningful comparison, we work on established neuroimaging datasets (where offline processing is targeted) and will publish all source code and datasets created in the process. 

We think that the presented view on hybrid cognition lead to advanced self-supervising artificial agents with emerging properties and provide a basis for intelligent and adaptive extended human cognition. We further suggest that the resulting hybrid generative models can expand our understanding of the human mind. With this in mind, we hope that this overview helps to make collaborations between researchers from fields such as machine learning, neuroscience and psychology more effective.

\newpage
\bibliography{references}

\end{document}